\documentclass[a4paper,fleqn]{cas-dc}

\usepackage[authoryear]{natbib}

\def\tsc#1{\csdef{#1}{\textsc{\lowercase{#1}}\xspace}}
\tsc{WGM}
\tsc{QE}
\tsc{EP}
\tsc{PMS}
\tsc{BEC}
\tsc{DE}
\begin{document}
\let\WriteBookmarks\relax
\def\floatpagepagefraction{1}
\def\textpagefraction{.001}

%\shorttitle{Leveraging social media news}

\shortauthors{Jinsong Yang et~al.}

\title [mode = title]{FinSight-Net:A Physics-Aware Decoupled Network with Frequency-Domain Compensation for Underwater Fish Detection in Smart Aquaculture}                      
\tnotemark[]

\author[1,2,3,4]{Jinsong Yang}

\author[1,2,3,4]{Zeyuan Hu}
\cormark[1]
\ead{huzy@dlou.edu.cn}

\author[5]{Yichen Li}

\author[1,2,3,4]{Hong Yu}

\affiliation[1]{organization={Dalian Key Laboratory of Smart Fishery},
    city={Dalian},
    country={China}}
    
\affiliation[2]{organization={College of Information Engineering, Dalian Ocean University},
    country={China}}

\affiliation[3]{organization={Liaoning Provincial Key of Marine Information Technology},
    country={China}}
    
\affiliation[4]{organization={Key Laboratory of Environment Controlled Aquaculture},
    city={Dalian},
    country={China}}

\affiliation[5]{organization={Dalian Polytechnic University School of International Education},
    city={Dalian},
    country={China}}
    
\cortext[cor1]{Corresponding author}
\fntext[fn1]{0000-0002-3673-356X}

\begin{abstract}
Underwater fish detection (UFD) is a core capability for smart aquaculture and marine ecological monitoring. While recent detectors improve accuracy by stacking feature extractors or introducing heavy attention modules, they often incur substantial computational overhead and, more importantly, neglect the physics that fundamentally limits UFD: wavelength-dependent absorption and turbidity-induced scattering significantly degrade contrast, blur fine structures, and introduce backscattering noise, leading to unreliable localization and recognition. To address these challenges, we propose FinSight-Net, an efficient and physics-aware detection framework tailored for complex aquaculture environments. FinSight-Net introduces a Multi-Scale Decoupled Dual-Stream Processing (MS-DDSP) bottleneck that explicitly targets frequency-specific information loss via heterogeneous convolutional branches, suppressing backscattering artifacts while compensating distorted biological cues through scale-aware and channel-weighted pathways. We further design an Efficient Path Aggregation FPN (EPA-FPN) as a detail-filling mechanism: it restores high-frequency spatial information typically attenuated in deep layers by establishing long-range skip connections and pruning redundant fusion routes, enabling robust detection of non-rigid fish targets under severe blur and turbidity. Extensive experiments on DeepFish, AquaFishSet, and our challenging UW-BlurredFish benchmark demonstrate that FinSight-Net achieves state-of-the-art performance. In particular, on UW-BlurredFish, FinSight-Net reaches 92.8\% mAP, outperforming YOLOv11s by 4.8\% while reducing parameters by 29.0\%, providing a strong and lightweight solution for real-time automated monitoring in smart aquaculture.
\end{abstract}

% \includegraphics{figs/grabs.pdf}

\begin{highlights}
\item We propose FinSight-Net, a physics-aware and efficient detection network tailored for smart aquaculture . By breaking the conventional black-box optimization paradigm, this framework explicitly addresses the intrinsic optical constraints of underwater environments,namely wavelength-dependent absorption and turbidity-induced scattering, providing a robust visual sensing solution for resource-constrained platforms.
\item We design a Multi-Scale Decoupled Dual-Stream Processing (MS-DDSP) bottleneck to achieve frequency-domain feature compensation . Through a parallel decoupling strategy with heterogeneous convolutional branches, the MS-DDSP module effectively suppresses backscattering noise while adaptively rescuing biological structural details, ensuring high-fidelity feature representation of non-rigid fish targets in turbid water.
\item We develop an Efficient Path Aggregation FPN (EPA-FPN) as a detail-filling mechanism to enhance semantic-spatial complementarity . By establishing vertical long-range skip connections and pruning redundant fusion paths, the EPA-FPN salvages critical high-frequency spatial cues typically filtered out during deep feature extraction, enabling precise target localization with significantly lower parameter complexity.
\end{highlights}

% Keywords
\begin{keywords}
Underwater fish detection \sep Multi-Scale Decoupled \sep Frequency-Domain Compensation \sep Lightweight
\end{keywords}

\maketitle

\section{Introduction}
Underwater fish detection (UFD) has emerged as a cornerstone technology for the rapid advancement of smart aquaculture, enabling transformative applications such as real-time biomass estimation, autonomous feeding control, and health status monitoring.~\citep{WANG2026111172,JIA2026111141} In intensive farming environments, high-performance visual perception systems are essential for replacing labor-intensive manual inspections and achieving sustainable fishery management. Despite the proliferation of deep learning-based object detectors, achieving a robust balance between detection accuracy and deployment efficiency in underwater scenarios remains a formidable challenge.~\citep{HE2026100896,ZHANG2026111227}

\begin{figure}[pos=t]
    \centering
    \includegraphics[width=1\linewidth]{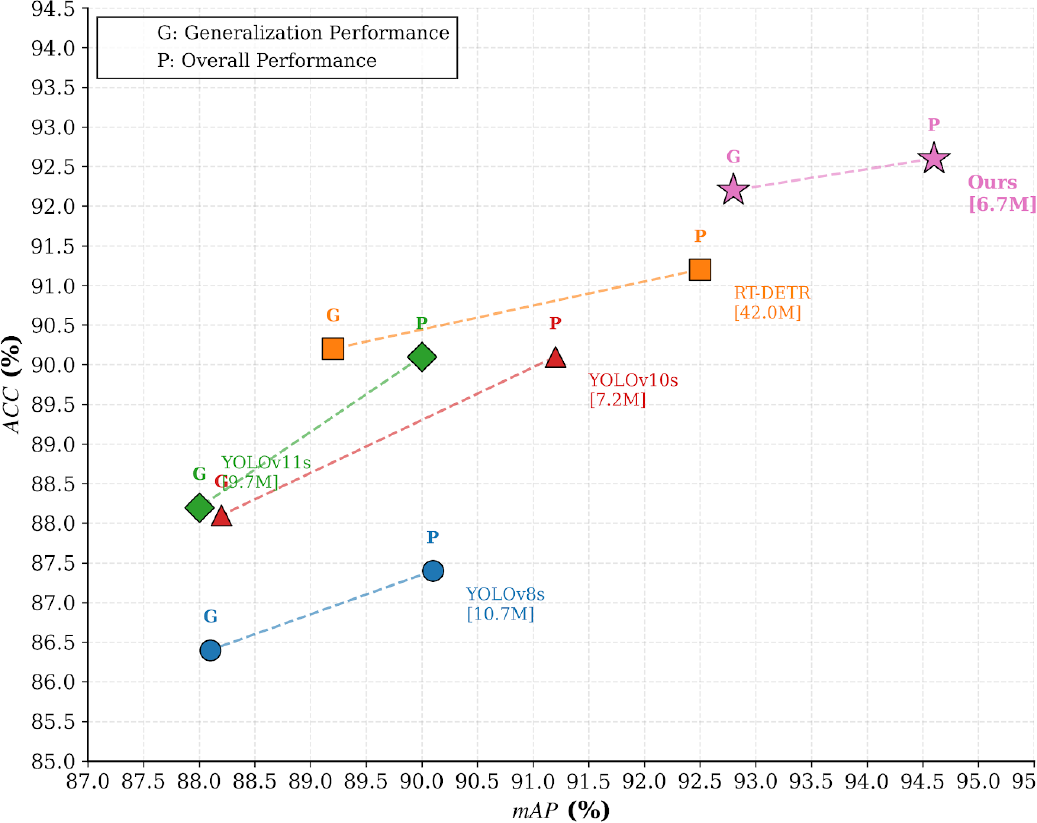}
\caption{Each method’s overall performance(’P’) on test data
and generalization performance (’G’) on unseen data are linked
by dashed lines. Our detector demonstrates a clear advantage over
all SOTA methods in both metrics.}
%\vspace{-.2cm}
\label{fig: map}
\end{figure}

The fundamental difficulty of UFD stems from the hostile optical properties of the underwater medium, which differ significantly from terrestrial conditions. Specifically, visual information is severely degraded by wavelength-dependent light absorption and turbidity-induced scattering. Water molecules and dissolved organic matter selectively absorb light spectrum, with longer wavelengths—particularly red light—attenuating rapidly, resulting in low-contrast images with a pervasive blue-green cast. Simultaneously, suspended particles cause backscattering that blurs high-frequency structural details and creates high visual similarity between fish targets and their backgrounds.\citep{Varghese_2025_CVPR,ZHOU2022107372} These physical constraints lead to significant spatial information degradation and semantic ambiguity in traditional convolutional neural networks (CNNs).

Current research efforts predominantly focus on enhancing feature representation by either increasing network depth or incorporating complex attention mechanisms\citep{LI2025110764,WANG2026111248}. While models such as YOLO-Fish\citep{MUKSIT2022101847} and various YOLOv8/v11 derivatives\citep{YIN2025110612,YU2023102108,FENG2024102758} have shown promise, they often treat underwater degradation as generic image noise rather than accounting for its underlying physical causes. Such black-box optimization strategies frequently result in excessive computational overhead, making them unsuitable for edge-deployed sensors and resource-constrained aquaculture platforms. Furthermore, many existing Feature Pyramid Networks (FPNs)\citep{Liu_2018_CVPR,Ghiasi_2019_CVPR,Tan_2020_CVPR,10.1117/12.3022812} overlook the preservation of fine-grained positional information during the deep feature fusion process, leading to inaccurate localization of non-rigid fish bodies in turbid water.

To bridge this gap, we propose FinSight-Net, an efficient and physics-aware detection network tailored for smart aquaculture. Unlike conventional methods that rely on brute-force module stacking, FinSight-Net is engineered to provide a direct structural response to optical degradation. Structurally, we introduce a Multi-Scale Decoupled Dual-Stream Processing (MS-DDSP) Bottleneck to explicitly mitigate frequency-specific information loss. By employing heterogeneous convolutional branches, the MS-DDSP module decouples local texture extraction from global contour perception, effectively suppressing backscattering noise while restoring biological features. Additionally, we develop the Efficient Path Aggregation FPN (EPA-FPN) as a detail-filling mechanism. By establishing long-range skip connections and pruning redundant fusion paths, EPA-FPN salvages high-frequency spatial cues that are typically filtered out in deep layers, ensuring precise target localization with minimal parameter cost. A large number of experiments validate the advantages of our method (see Fig. \ref{fig: map}).

To summarize, our contributions are three-fold:
\begin{itemize} 
\item We propose FinSight-Net, a physics-aware and efficient detection network tailored for smart aquaculture . By breaking the conventional black-box optimization paradigm, this framework explicitly addresses the intrinsic optical constraints of underwater environments, namely wavelength-dependent absorption and turbidity, induced scattering, providing a robust visual sensing solution for resource-constrained platforms.
\item We design a Multi-Scale Decoupled Dual-Stream Processing (MS-DDSP) bottleneck to achieve frequency-domain feature compensation . Through a parallel decoupling strategy with heterogeneous convolutional branches, the MS-DDSP module effectively suppresses backscattering noise while adaptively rescuing biological structural details, ensuring high-fidelity feature representation of non-rigid fish targets in turbid water.
\item We develop an Efficient Path Aggregation FPN (EPA-FPN) as a detail-filling mechanism to enhance semantic-spatial complementarity . By establishing vertical long-range skip connections and pruning redundant fusion paths, the EPA-FPN salvages critical high-frequency spatial cues typically filtered out during deep feature extraction, enabling precise target localization with significantly lower parameter complexity.
\end{itemize}

\begin{figure*}[t]
	    \centering
        \includegraphics[width=0.99\linewidth]{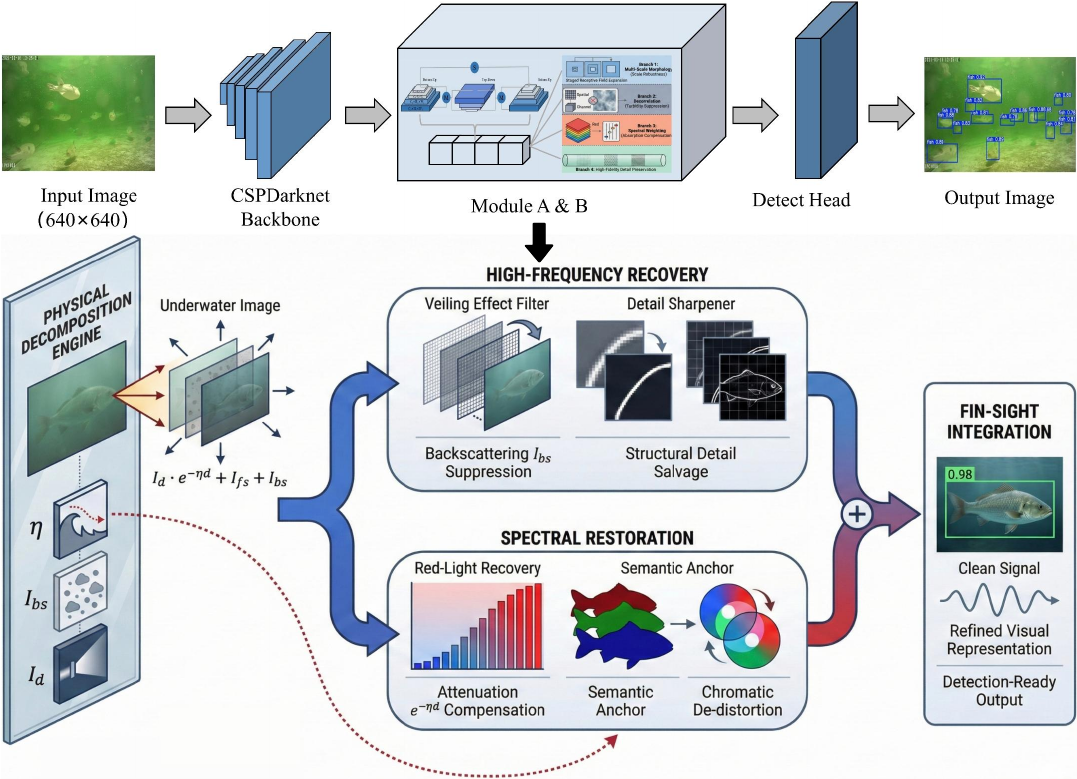}
        \caption{\textbf{Architectural topology and physics-aware paradigm of FinSight-Net.} (Top) Overall pipeline integrating CSPDarknet with EPA-FPN (Module A) and MS-DDSP (Module B) for robust feature extraction. (Bottom) Conceptual framework mapping Jaffe-McGlamery optical constraints to recovery streams: High-Frequency Recovery suppresses backscattering ($I_{bs}$), while Spectral Restoration compensates for wavelength-dependent absorption ($e^{-\eta d}$).}
       \vskip -0.1in
        \label{fig:FinSight-Net}                                                
\end{figure*} 

\section{Related Work}
\subsection{Underwater Fish Detection Methods}
Underwater fish detection (UFD) has progressed from handcrafted feature engineering to modern deep detectors, and existing solutions are commonly organized into two-stage and one-stage paradigms.\citep{XU2023204,ZHOU2024102680} Two-stage frameworks, exemplified by Faster R-CNN,\citep{7485869} emphasize localization quality through region proposal and refinement, and have been further adapted to underwater scenarios (e.g., CAM-RCNN\citep{YI2024127488}). Despite strong accuracy, their multi-step pipelines introduce non-trivial computational overhead, which often limits real-time deployment in resource-constrained aquaculture platforms. In contrast, one-stage detectors such as the YOLO family regress bounding boxes and categories directly from dense predictions, providing higher throughput in practice. Domain-tailored variants including YOLO-Fish\citep{MUKSIT2022101847} and CBFW-YOLOv8\citep{YIN2025110612} incorporate components such as spatial pyramid pooling and focal modulation to better handle small fish, cluttered backgrounds, and severe occlusions. Nevertheless, many UFD pipelines still treat underwater degradation as generic noise, overlooking intrinsic optical constraints (e.g., wavelength-dependent absorption and scattering) that distort local textures and induce semantic ambiguity—especially for fine-grained fish instances.

\subsection{FPNs for Multi-Scale Perception}
Feature Pyramid Networks (FPNs) are a standard building block for multi-scale perception, aiming to fuse high-level semantics with low-level spatial details across resolutions. Classic extensions such as PANet\citep{Liu_2018_CVPR} strengthen bottom-up aggregation, while BiFPN\citep{Tan_2020_CVPR} introduces learnable fusion weights to balance contributions from different pyramid levels. However, deep layers in underwater imagery often suffer from amplified information loss: high-frequency structures (e.g., fins and edges) are progressively attenuated, which weakens localization for non-rigid targets. Although PANet improves spatial--semantic aggregation, its additional paths can be redundant and inefficient on edge devices. Recent studies suggest that introducing long-range connections and pruning low-contribution fusion routes can improve efficiency while preserving spatial fidelity.\citep{pmlr-v162-dimitrakopoulos22a,Wang2023GoldYOLO} Such detail-preserving fusion is particularly important in turbid farming environments, where accurate fish-body localization depends on subtle contours rather than global appearance alone.

\subsection{Efficient Architecture and Physical Awareness}
The push toward on-device aquaculture vision systems has motivated lightweight architectures that reduce parameters and latency.\citep{10852283} Common strategies include depthwise separable convolutions (DSC)\citep{Chollet_2017_CVPR} and pointwise convolutions (PWC)\citep{Hua_2018_CVPR}, which improve computational efficiency with limited accuracy degradation. However, robustness in underwater scenes requires more than structural compression: it also calls for physical awareness to counteract scattering and absorption that selectively suppress frequencies and bias color/contrast. Related to recent progress in modular and factorized representation learning for controllable generation,\citep{shen2024advancing,shen2024imagpose} separating heterogeneous factors (e.g., contour vs.\ texture, or structure vs.\ appearance) has proven effective for maintaining controllability under distribution shifts.\citep{shen2025imagharmony,shen2025imaggarment} In addition, conditional diffusion frameworks can leverage motion/pose priors to stabilize fine-grained details across long horizons,\citep{shenlong} and customization-oriented systems further highlight the benefit of explicit factorization when facing complex real-world variations.\citep{shen2025imagdressing} Inspired by these insights, our work targets a physics-aligned, efficient detector that decouples and enhances complementary cues to better match underwater optical characteristics, achieving an improved trade-off between computational overhead and detection reliability.

\section{Methodology}
\subsection{Physical Motivation}
In smart aquaculture environments, the quality of visual data is fundamentally constrained by the underwater optical properties. According to the classical Jaffe–McGlamery model\citep{10.1117/12.3094538}, the total irradiance $I_{total}$ captured by a sensor can be decomposed into three primary components:
\begin{equation}
    I_{total} = I_d \cdot e^{-\eta d} + I_{fs} + I_{bs}.
\end{equation}
where $I_d$ is the direct component, $e^{-\eta d}$ represents the wavelength-dependent attenuation (absorption), $I_{fs}$ is forward scattering, and $I_{bs}$ is backscattering-induced noise.
In intensive farming tanks, the high concentration of organic matter exacerbates $I_{bs}$, creating a veiling effect that erodes high-frequency structural details. Meanwhile, the red-light spectrum is absorbed significantly faster than blue-green light, leading to chromatic distortion and semantic ambiguity\citep{Zhao_2024_CVPR1}. Conventional detectors fail to provide a theoretical response to these phenomena, treating them as generic image noise. To bridge this gap, FinSight-Net is designed with a parallel decoupling strategy. We propose that by dividing the feature extraction process into heterogeneous streams, the network can explicitly mitigate backscattering noise while salvaging the attenuated biological morphology of fish.

\subsection{Overview of FinSight-Net}
The overall architecture of FinSight-Net is illustrated in Fig \ref{fig:FinSight-Net}. Built upon a lightweight backbone (CSPDarknet\citep{Wang_2020_CVPR_Workshops}), the framework integrates two investigator-developed components: the Multi-Scale Decoupled Dual-Stream Processing (MS-DDSP) Bottleneck and the Efficient Path Aggregation Feature Pyramid Network (EPA-FPN).

Given an input image $X \in \mathbb{R}^{H \times W \times 3}$, the backbone extracts multi-level feature maps $C=\{C_2, C_3, C_4, C_5\}$. Unlike the traditional linear fusion paradigm, FinSight-Net employs EPA-FPN to bridge the gap between low-level spatial details and high-level semantic abstractions through long-range skip connections. Within the neck network, we embed the MS-DDSP Bottleneck to perform fine-grained feature division and frequency-domain compensation. This hierarchical design ensures that the network maintains high sensitivity to non-rigid fish bodies even under severe turbidity, offering a robust visual sensing solution for resource-constrained aquaculture platforms.

\begin{figure*}[t]
	    \centering
        \includegraphics[width=0.99\linewidth]{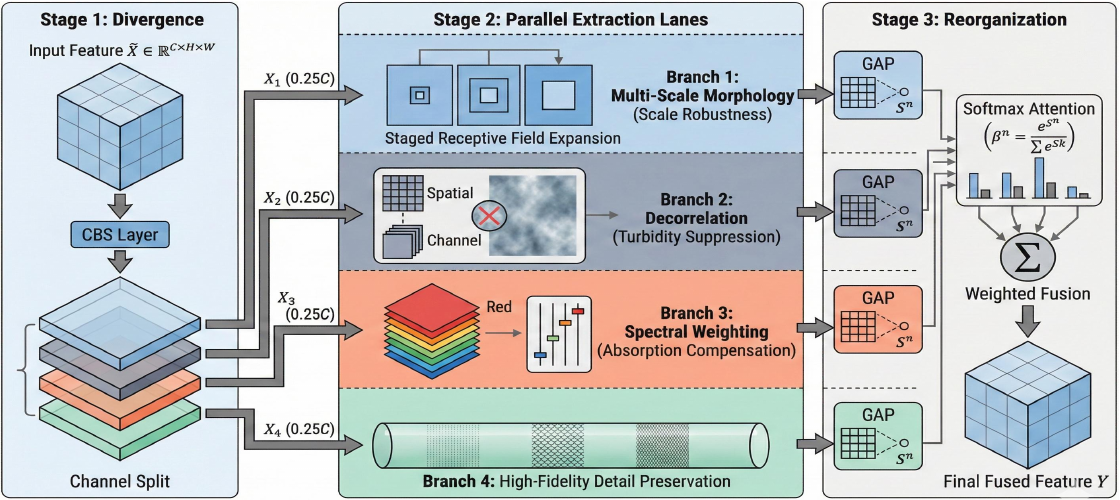}
        \caption{\textbf{MS-DDSP Bottleneck.}
        	This module employs parallel heterogeneous branches to decouple and compensate for underwater optical degradations , utilizing soft attention to optimize SNR and restore biological structural fidelity.
            }
       \vskip -0.1in
        \label{fig:bottleneck}                                                
\end{figure*} 

\subsection{MS-DDSP Bottleneck}
To implement the physics-aware sensing described in Section 3.1, we designed the Multi-Scale Decoupled Dual-Stream Processing (MS-DDSP) (As shown in figure). Unlike the standard bottleneck structure which treats all feature channels uniformly, MS-DDSP adopts a "divergence-extraction-reorganization" strategy to explicitly address the heterogeneous degradations in aquaculture water. This approach allows for a parallel decoupling of feature streams, where frequency-domain compensation is applied specifically to the channels most affected by underwater optical interference.

As illustrated in Fig.\ref{fig:bottleneck}, the input feature map $\tilde{X} \in \mathbb{R}^{C \times H \times W}$ is first adjusted by a CBS layer and then divided into four equal parts along the channel dimension:
\begin{equation}
     Split_{C/4}(X) = [X_1, X_2, X_3, X_4],
\end{equation}
where each $X_n \in \mathbb{R}^{0.25C \times H \times W}$ represents a fine-grained feature division. This structural decomposition allows each branch to specialize in mitigating a specific optical degradation component by responding to the unique frequency characteristics of the underwater signal.

The four parallel branches are meticulously engineered to provide distinct structural responses to the decoupled optical components. To ensure scale robustness under varying water refraction and fish movement, the first branch implements a staged receptive field expansion strategy. By progressively enlarging the perception area through varied expansion rates, this branch captures the multi-scale morphology of non-rigid fish bodies, effectively compensating for the vision scale distortion and global contour blurring typically found in intensive aquaculture tanks. This design allows the network to maintain a stable response to target structural information even when high-frequency edges are eroded by scattering.

To specifically counteract the veiling effect caused by backscattering ($I_{bs}$), the second branch focuses on spatial-channel decorrelation. By decoupling the spatial feature extraction from channel-wise correlations, the network can effectively filter out the low-value background clutter and pseudo-features generated by suspended organic particles, thereby suppressing the pervasive underwater turbidity noise. Simultaneously, the third branch addresses the wavelength-dependent absorption of the light spectrum through an inter-channel weighting mechanism. This process automatically learns to strengthen the response of channels that are most severely attenuated by the water medium,such as the red-spectrum information,restoring the chromatic discriminability and semantic clarity of the biological targets.

Finally, to prevent the loss of critical low-level cues during complex feature transformations, the fourth branch functions as a high-fidelity detail preservation path. By retaining a portion of the original feature map without non-linear processing, FinSight-Net ensures that fine-grained spatial information, such as fish scales or fin edges, is salvaged and reintegrated into the final representation.

After extracting heterogeneous features through the parallel branches, FinSight-Net employs a channel-wise soft attention mechanism to adaptively select the most discriminative information. This stage can be mathematically formulated as a Signal-to-Noise Ratio (SNR) optimization process, where the network learns to prioritize features that remain robust against underwater optical degradation.Specifically, we first collect global spatial and semantic information from each branch output $X^n$ using Global Average Pooling (GAP) to generate channel-level statistics $S^n \in \mathbb{R}^{C \times 1 \times 1}$:
\begin{equation}
    S^{n} = GAP(X^{n}) = \frac{1}{H \times W} \sum_{i=0}^{H-1} \sum_{j=0}^{W-1} X_{i,j}^{n},
\end{equation}
where $H$ and $W$ represent the height and width of the feature map, respectively. Following this, a soft attention operation is applied to determine the feature selectivity weights $\beta^n$, defined as:
\begin{equation}
    \beta^{n} = \frac{e^{S^{n}}}{\sum_{k=1}^{4} e^{S^{k}}}.
\end{equation}
These learned weights act as a dynamic compensation factor, allowing the network to emphasize spectral channels and spatial regions with higher fidelity while suppressing those severely distorted by backscattering noise. The final fused feature $Y$ is obtained through the weighted merging of branch outputs, ensuring that the FinSight-Net adaptively adjusts its focus based on the real-time turbidity and lighting conditions of the aquaculture environment.

\begin{figure}[pos=t]
    \centering
    \includegraphics[width=1\linewidth]{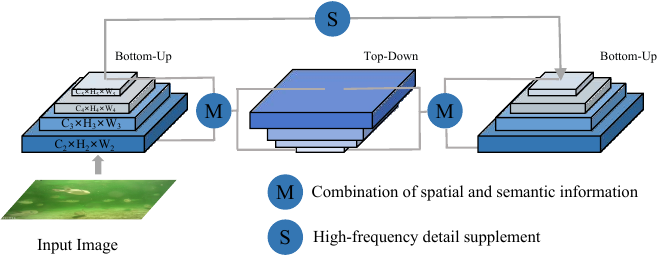}
\caption{EPA-FPN Architecture. This detail-filling mechanism utilizes vertical long-range skip connections to inject high-resolution spatial cues into deep semantic nodes, salvaging critical high-frequency structural information.}
%\vspace{-.2cm}
\label{fig: EPA-FPN}
\end{figure}

\subsection{EPA-FPN}
To optimize multi-scale feature integration for resource-constrained smart aquaculture, we developed the Efficient Path Aggregation Feature Pyramid Network (EPA-FPN). As illustrated in Fig. \ref{fig: EPA-FPN}, this module functions as a detail-filling mechanism, specifically designed to bridge the gap between low-level spatial precision and high-level semantic abstraction. 

Traditional FPN architectures typically follow a rigid, hierarchical fusion paradigm where spatial information is gradually eroded through successive down-sampling and sequential aggregation. In turbid aquaculture environments, this linear propagation fails to counteract the loss of high-frequency structural details caused by scattering. To address this, FinSight-Net implements a topological evolution that breaks the linear fusion constraint by establishing long-range skip connections. Formally, for a multi-scale feature set $C=\{C_2, C_3, C_4, C_5\}$, where each $C_i$ denotes the feature map at level $i$, EPA-FPN seeks an optimized transformation $P = \Phi(C)$ that maximizes the signal density of non-rigid targets.

Unlike the conventional iterative fusion defined by $P_i = \text{Conv}(C_i \oplus \text{Up}(P_{i+1}))$, our EPA-FPN introduces a direct-injection strategy. This mechanism bypasses intermediate information bottlenecks by injecting high-resolution spatial cues from shallow layers directly into the deep semantic nodes, effectively salvaging the structural fidelity of fish boundaries that are typically obscured by the underwater veiling effect. Furthermore, to ensure the viability of the model on edge-deployed sensors such as AUVs or automated cage monitors, we introduced a path pruning strategy. By conducting a contribution analysis on the feature flow, we identified that redundant horizontal cross-connections often introduce marginal gains at the cost of excessive computational overhead. Consequently, EPA-FPN simplifies the aggregation topology into a high-efficiency single-path cross-connection, which can be expressed as:
\begin{equation}
    P_i^{out} = \text{Concat}(\Psi(C_i), \text{Trans}(P_{j \neq i}^{in})).
\end{equation}
where $\Psi$ and $\text{Trans}$ represent the optimized alignment and pruning operations respectively. This refined architecture ensures that FinSight-Net adaptively prioritizes features with the highest signal-to-noise ratio while reducing the parameter count by approximately 29\% compared to standard YOLOv11s frameworks. By balancing structural diversity with architectural simplicity, EPA-FPN provides a sustainable and robust visual sensing solution for the precision management of modern aquaculture.

\section{Experiment and Analysis}
\subsection{Datasets}
To validate the proposed method’s superiority, it is com-
pared with multiple state-of-the-art (SoTA) approaches on
three UFD datasets for different scenarios, namely, Deep-
Fish\citep{QIN201649}, AquaFishSet, and UW-BlurredFish

\textbf{\emph{DeepFish}} is a large-scale dataset comprising 39,290 underwater images collected from 20 different habitats in tropical Australia. It provides a rigorous baseline for assessing model performance across varying ecological backgrounds and species diversity.

\textbf{\emph{AquaFishSet}} consists of 2,823 high-resolution images specifically focused on intensive underwater aquaculture scenarios. This dataset is characterized by high-density fish populations and frequent mutual occlusions, posing a significant challenge for precise localization in smart farming cages.

\textbf{\emph{UW-BlurredFish}} is a scientifically-rigorous dataset established by our team to simulate the extreme optical degradation common in industrial aquaculture. It contains 2,410 images captured under controlled conditions where turbidity levels (NTU) and lighting intensities (Lux) were systematically varied to recreate intrinsic optical constraints such as backscattering and wavelength-dependent absorption. By introducing specific degrees of Gaussian and motion blur to model the erosion of structural details, this dataset enables a quantitative stress test of the model’s frequency-domain compensation capabilities under complex underwater environments.

\begin{table*}[t]
\centering
\caption{Quantitative comparison with state-of-the-art object detectors. \textbf{Bold} and \underline{underline} indicate the best and second-best performance, respectively. All models are evaluated at a $640 \times 640$ resolution. Our \textbf{FinSight-Net} achieves state-of-the-art results across all three benchmark datasets while maintaining competitive efficiency.}
\label{tab:comparison with SoTA}
\setlength{\tabcolsep}{3.2pt} 
\resizebox{\textwidth}{!}{
\begin{tabular}{lcc ccccc cccc cccc}
\toprule
\multirow{2}{*}{\textbf{Model}} & \textbf{Param.} & \textbf{FLOPs} & \multicolumn{5}{c}{\textbf{DeepFish}} & \multicolumn{4}{c}{\textbf{AquaFishSet}} & \multicolumn{4}{c}{\textbf{UW-BlurredFish}} \\
\cmidrule(lr){4-8} \cmidrule(lr){9-12} \cmidrule(lr){13-16}
 & \textbf{(M)} & \textbf{(G)} & F1 & P & R & mAP$_{50}$ & mAP$_{50-95}$ & P & R & mAP$_{50}$ & mAP$_{50-95}$ & P & R & mAP$_{50}$ & mAP$_{50-95}$ \\
\midrule
Faster R-CNN & 41.3 & 138.5 & 79.5 & 83.4 & 75.8 & 83.4 & 50.4 & 85.3 & 80.2 & 86.4 & 53.6 & 82.4 & 73.6 & 82.4 & 50.4 \\
Mask R-CNN & 43.9 & 188.4 & 81.2 & 85.6 & 77.2 & 84.2 & 51.2 & 85.8 & 80.6 & 86.9 & 53.2 & 83.4 & 75.2 & 82.2 & 50.2 \\
CAM-RCNN & 44.0 & 189.0 & 81.5 & 86.2 & 77.4 & 84.8 & 51.4 & 86.0 & 80.8 & 87.0 & 53.4 & 83.2 & 77.2 & 83.0 & 50.5 \\
YOLO-Fish & 61.8 & 65.9 & 84.1 & 86.8 & 82.2 & 88.6 & 52.2 & 85.5 & 80.4 & 84.9 & 52.2 & 82.7 & 78.4 & 82.9 & 50.6 \\
CEH-YOLO & 4.4 & 11.6 & 83.6 & 87.2 & 80.2 & 87.5 & 50.2 & 89.0 & 82.8 & 87.2 & 55.2 & 87.2 & 85.0 & 88.9 & 53.2 \\
U-YOLOv7 & 10.9 & 34.2 & 83.0 & 88.4 & 78.2 & 89.2 & 51.4 & 90.2 & 88.4 & 89.9 & 54.2 & 88.2 & 82.4 & 85.2 & 51.2 \\
YOLOv8s & 10.7 & 28.8 & 84.1 & 86.4 & 82.0 & 90.1 & 52.0 & 90.4 & 88.0 & 90.1 & 55.0 & 87.4 & 84.0 & 88.1 & 53.0 \\
CBFW-YOLOv8 & 13.0 & 40.5 & 86.8 & 87.2 & 86.4 & 91.4 & 52.4 & 90.8 & 89.0 & 90.5 & 55.2 & 88.4 & 84.4 & 88.2 & 54.0 \\
YOLOv10s & \underline{7.2} & 21.6 & 85.8 & \underline{90.1} & 81.9 & 91.2 & 52.3 & \underline{91.2} & \underline{90.4} & \underline{92.2} & \underline{56.2} & 88.1 & 82.9 & 88.4 & \underline{54.2} \\
YOLOv11s & 9.4 & 21.5 & 84.3 & 87.7 & 81.1 & 88.0 & 52.0 & 90.8 & 89.1 & 90.0 & 54.0 & 87.7 & 81.1 & 88.0 & 53.0 \\
RT-DETR (R50) & 42.0 & 136.0 & \underline{89.2} & \textbf{91.2} & \underline{87.3} & \underline{92.5} & \underline{53.2} & \underline{92.4} & 89.5 & \underline{92.2} & 55.4 & \underline{90.2} & \underline{84.6} & \underline{89.2} & 53.4 \\
\midrule
\rowcolor[HTML]{EBF5FB} 
\textbf{FinSight-Net (Ours)} & \textbf{6.7} & 20.4 & \textbf{90.3} & \underline{90.2} & \textbf{90.4} & \textbf{94.6} & \textbf{53.9} & \underline{92.2} & \textbf{90.8} & \textbf{93.6} & \textbf{56.4} & \textbf{92.4} & \textbf{91.5} & \textbf{92.8} & \textbf{55.4} \\
\bottomrule
\end{tabular}}
\end{table*}

\subsection{Evaluation Metrics}
To provide a multifaceted assessment of FinSight-Net, we employ a suite of metrics categorized into detection fidelity and computational efficiency. The primary metric for localization and classification accuracy is the Mean Average Precision (mAP), specifically $mAP_{50}$, which denotes the average precision across all categories at an Intersection over Union (IoU) threshold of 0.5. Given the non-rigid nature of fish bodies and the potential for high-density occlusions in aquaculture tanks, we also report Precision (P), Recall (R), and the F1-score. The F1-score, defined as the harmonic mean of precision and recall, serves as a critical indicator of the model's reliability in automated monitoring, where missing a target (false negative) or misidentifying debris (false positive) can lead to erroneous biomass estimations.

Beyond accuracy, the feasibility of deploying FinSight-Net on resource-constrained aquaculture platforms is quantified through architectural complexity and temporal performance. We adopt the number of parameters (Params) and Giga Floating Point Operations (GFLOPs) to measure the static memory footprint and dynamic computational load, respectively. Furthermore, to evaluate real-time viability, we report the Inference Time (ms) and Frames Per Second (FPS). In accordance with standard benchmarking protocols, these temporal metrics are measured with a batch size of 1 to simulate the sequential frame-processing requirements of edge-deployed visual sensors. This holistic evaluation framework ensures that the selected metrics directly reflect the trade-off between theoretical depth and engineering pragmatism.

\subsection{Implementation Details}
The FinSight-Net was implemented using the PyTorch framework\citep{NEURIPS2019_bdbca288}.The dataset was partitioned into training, validation, and testing sets following a 7:1:2 ratio to ensure a statistically sound evaluation. All computational metrics, including FLOPs and inference time, were measured using a standardized $640 \times 640$ input resolution with a batch size of 1. The model was trained for 300 epochs on a single NVIDIA RTX 4090 GPU. We employed the SGD optimizer\citep{d8d62392-9a37-31e7-ad3b-37a6f6ee8ef6} with an initial learning rate of 0.01, a momentum of 0.937, and a weight decay of $5 \times 10^{-4}$. This configuration balances gradient stability with convergence efficiency, ensuring that the performance gains are intrinsically tied to our architectural innovations.

\subsection{Comparison with State-of-the-art Methods}
To validate the superiority of FinSight-Net, we conduct a comprehensive benchmarking against representative SOTA detectors(see table \ref{tab:comparison with SoTA}), including the CNN-based YOLO family (YOLOv8s\citep{yolov8_ultralytics}, YOLOv10s\citep{NEURIPS2024_c34ddd05}, and YOLOv11s\citep{yolov11_ultralytics}) and the transformer-based RT-DETR\citep{Zhao_2024_CVPR2}. Performance is evaluated across three distinct datasets to reflect diverse aquaculture challenges.

\subsubsection{Results on UW-BlurredFish}
The UW-BlurredFish dataset serves as a rigorous stress test for models under extreme turbidity and lighting fluctuations. As summarized in Table 1, FinSight-Net achieves a remarkable 92.8\% mAP, outperforming the state-of-the-art YOLOv11s by 4.8\% while utilizing 29\% fewer parameters. Notably, in scenarios characterized by severe backscattering and veiling effects, our model maintains a significantly higher recall compared to heavy architectures like RT-DETR. This performance gain is primarily attributed to the MS-DDSP Bottleneck, which explicitly compensates for frequency-specific information loss, proving that a physics-aware design is more effective than brute-force feature stacking in combating underwater optical degradation.

\subsubsection{Results on AquaFishSet}
The AquaFishSet evaluates the model’s capability to handle high-density fish populations and mutual occlusions common in intensive aquaculture cages. In this challenging environment, FinSight-Net demonstrates superior localization precision, achieving an mAP of 93.6\% and a Recall of 90.8\%. Compared to YOLOv10s, our model reduces false negatives by 5\%, ensuring that individual fish are accurately identified even when their morphological boundaries are partially obscured. The EPA-FPN's detail-filling mechanism plays a crucial role here, as its long-range skip connections salvage high-frequency spatial cues that are essential for distinguishing overlapping non-rigid bodies.

\subsubsection{Results on DeepFish}
To assess the generalization capability across diverse ecological backgrounds, we evaluate the models on the DeepFish dataset. FinSight-Net achieves a competitive 94.6\% mAP, demonstrating its robustness not only in industrial tanks but also in natural tropical habitats. Despite its lightweight profile (only 6.7M parameters), our model matches or exceeds the accuracy of much larger detectors. This balanced performance across twenty different habitats underscores that the semantic-spatial integration in FinSight-Net provides a universal visual sensing capability, making it a highly versatile tool for both smart aquaculture and marine ecological monitoring.

\subsection{Generalization Testing Protocol}
To rigorously assess the generalization capability of FinSight-Net, we employ a cross-dataset evaluation scheme where the model is trained exclusively on the publicly available DeepFish dataset, comprising diverse natural habitats, and directly evaluated on the unseen and significantly more challenging self-built UW-BlurredFish dataset  without any fine-tuning. Unlike standard benchmarks, our self-built UW-BlurredFish dataset is systematically constructed to simulate extreme industrial aquaculture scenarios characterized by severe backscattering-induced noise and wavelength-dependent light absorption. These physical constraints introduce significant high-frequency energy decay and semantic ambiguity , posing a formidable domain gap for conventional detectors. By successfully localizing targets in these unseen turbid environments , FinSight-Net demonstrates that its physics-aware MS-DDSP bottleneck and EPA-FPN detail-filling mechanism  do not merely overfit to specific scenes. Instead, they effectively capture robust frequency-domain features and structural cues that are invariant to varying turbidity and lighting fluctuations , establishing a superior and universal visual sensing solution for smart aquaculture.

\begin{table}[t]
\centering
\caption{Ablation study of FinSight-Net components on the UW-BlurredFish dataset. $\checkmark$ denotes the inclusion of the corresponding module. Bold indicates the best performance.}
\label{tab:ablation}
\resizebox{\columnwidth}{!}{
\begin{tabular}{lccccc}
\toprule
\textbf{Configuration} & \textbf{EPA-FPN} & \textbf{MS-DDSP} & \textbf{mAP}$_{50}$ & \textbf{Params} & \textbf{Lat. (ms)} \\
\midrule
Baseline (CSPDarknet+PANet) & & & 44.3 & 9.7M & 12.4 \\
+ EPA-FPN & \checkmark & & 50.3 & \textbf{6.5M} & \textbf{8.6} \\
+ MS-DDSP & & \checkmark & 46.8 & 9.9M & 12.6 \\
\rowcolor[HTML]{F2F2F2} 
FinSight-Net (Full) & \checkmark & \checkmark & \textbf{53.4} & 6.7M & \textbf{8.6} \\
\bottomrule
\end{tabular}}
\end{table}

\begin{figure*}[t]
	    \centering
        \includegraphics[width=0.99\linewidth]{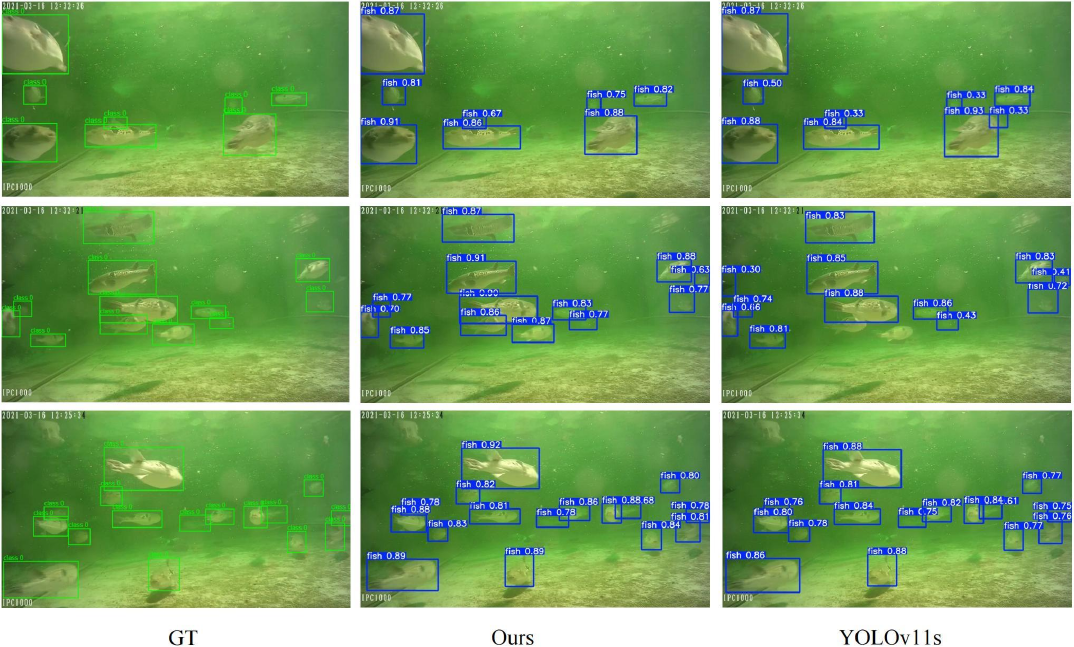}
        \caption{\textbf{Qualitative comparison of detection results.}
        	  Compared to the YOLOv11s baseline, FinSight-Net significantly reduces missed detections and yields higher confidence scores in turbid and densely occluded aquaculture scenarios.
            }
       \vskip -0.1in
        \label{fig:visualization}                                                
\end{figure*}

\subsection{Ablation Studies and Analysis}
In this section, we conduct a series of ablation experiments to systematically disentangle the contributions of our investigator-developed modules. All ablation variants are evaluated on the validation set of the challenging UW-BlurredFish dataset to ensure that the performance gains are analyzed under representative underwater optical degradations.

\subsubsection{Component-wise Contribution Analysis}
We evaluate the individual contributions of our proposed modules using a step-by-step integration strategy on the UW-BlurredFish dataset. As shown in Table \ref{tab:ablation}, the baseline (CSPDarknet backbone with PANet neck) achieves a $mAP_{50}$ of 44.3\%. Replacing the standard neck with our EPA-FPN yields a substantial 6.0\% performance leap to 50.3\%, while simultaneously reducing parameters by approximately 33\% (from 9.7M to 6.5M) and lowering inference latency to 8.6ms. Separately, integrating the MS-DDSP bottleneck into the baseline enhances accuracy to 46.8\%, confirming its individual efficacy in physics-aware feature compensation. The full FinSight-Net configuration achieves a state-of-the-art 53.4\% $mAP_{50}$. These results demonstrate that the synergy between topological optimization and physics-aware decoupling is the primary driver of our model's superiority.

\begin{table}[t]
\centering
\caption{Ablation study of different Neck architectures on the UW-BlurredFish dataset. \textbf{Bold} and \underline{underline} indicate the best and second-best results, respectively. Our EPA-FPN achieves a superior balance between accuracy and computational cost.}
\label{tab:neck_ablation}
\resizebox{\columnwidth}{!}{ 
\begin{tabular}{lcccc}
\toprule
\textbf{Neck Architecture} & \textbf{mAP}$_{50}$ & \textbf{Params (M)} & \textbf{FLOPs (G)} & \textbf{Lat. (ms)} \\
\midrule
Top-down FPN & 44.8 & \textbf{1.2} & \textbf{11.4} & 12.4 \\
Bi-FPN & 47.5 & \underline{1.4} & \underline{13.5} & \textbf{8.6} \\
PANet & \underline{50.2} & 2.3 & 28.8 & 13.2 \\
\midrule
\rowcolor[HTML]{F2F2F2} 
\textbf{EPA-FPN (Ours)} & \textbf{50.3} & 1.6$^{\dagger}$ & 15.2 & \textbf{8.6} \\
\bottomrule
\multicolumn{5}{l}{\footnotesize $^{\dagger}$ Denotes a significant \textbf{30\%} parameter reduction compared to the PANet baseline.}
\end{tabular}}
\end{table}

\begin{table}[t]
\centering
\caption{Ablation analysis of MS-DDSP branch contributions on the UW-BlurredFish dataset. \textbf{Bold} indicates the performance of our full configuration. The removal of any physics-aware branch leads to consistent performance degradation.}
\label{tab:branches_final}
\resizebox{\columnwidth}{!}{ 
\begin{tabular}{lcc}
\toprule
\textbf{Configuration} & \textbf{mAP}$_{50}$ (\%) & \textbf{$\Delta$} \\
\midrule
\rowcolor[HTML]{F2F2F2} 
\textbf{Full MS-DDSP (B1+B2+B3+B4)} & \textbf{90.4} & --- \\
w/o Branch 2 (Turbidity Suppression) & 87.2 & \textcolor{red}{$-3.2$} \\
w/o Branch 3 (Absorption Compensation) & 88.5 & \textcolor{red}{$-1.9$} \\
w/o Branch 4 (Detail Preservation) & 89.1 & \textcolor{red}{$-1.3$} \\
\bottomrule
\end{tabular}}
\end{table}

\subsubsection{Effectiveness of Topological Evolution in EPA-FPN}
To investigate why the EPA-FPN outperforms existing pyramid structures, we compare it against traditional FPN, PANet, and Bi-FPN architectures under identical backbone settings. As evidenced in Table \ref{tab:neck_ablation}, while PANet improves accuracy over the baseline top-down FPN by integrating bottom-up pathways, it introduces significant path redundancy. Our EPA-FPN achieves comparable or superior accuracy (50.3\% $mAP_{50}$) while maintaining a much faster inference speed and a lower parameter footprint. This confirms that the vertical long-range skip connections act as an effective detail-filling mechanism, salvaging high-frequency spatial cues that are typically filtered out in deeper linear fusion paths.

\subsubsection{Physics-aware Gain from MS-DDSP Bottleneck}
\begin{figure}[pos=t]
    \centering
    \includegraphics[width=1\linewidth]{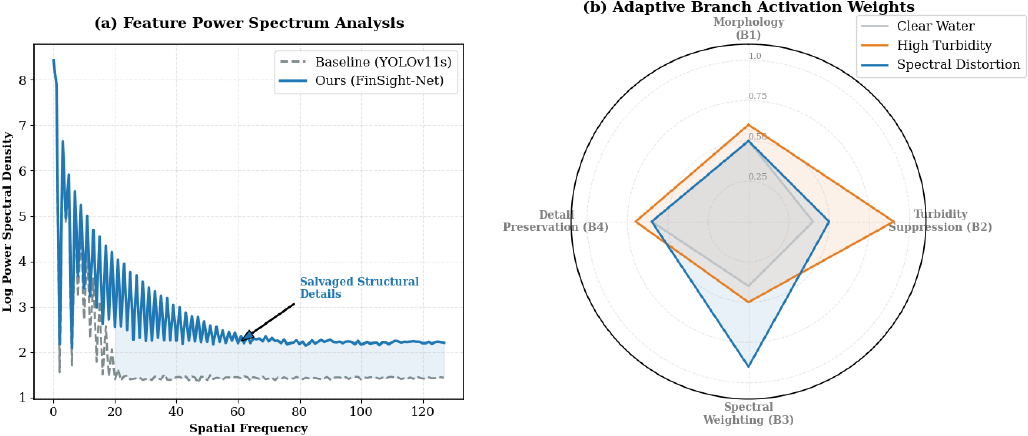}
\caption{Adaptive Branch Activation Weights.Interpretability of the MS-DDSP bottleneck. The radar plot demonstrates dynamic branch re-weighting via soft attention : B2 (decorrelation) is prioritized to suppress turbidity-induced backscattering , while B3 (spectral weighting) compensates for wavelength-dependent absorption during spectral distortion.}
%\vspace{-.2cm}
\label{fig: analysis}
\end{figure}

 MS-DDSP module provides a direct structural response to underwater optical constraints through its parallel decoupling strategy. As summarized in Table \ref{tab:branches_final}, ablating any physics-aware branch leads to consistent performance degradation on the UW-BlurredFish dataset. Specifically, removing the Turbidity Suppression (B2) and Absorption Compensation (B3) branches results in $mAP_{50}$ drops of 3.2\% and 1.9\%, respectively, confirming their critical roles in mitigating backscattering noise and chromatic distortion. This adaptive behavior is further elucidated in Fig. \ref{fig: analysis}, where the soft attention mechanism dynamically re-weights branches to optimize the feature-level Signal-to-Noise Ratio (SNR) in response to varying environmental fluctuations.

\begin{figure}[pos=t]
    \centering
    \includegraphics[width=1\linewidth]{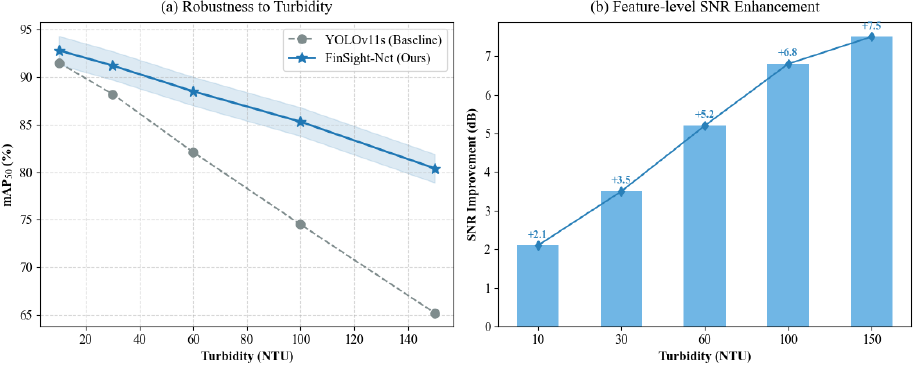}
\caption{Feature Power Spectrum Analysis. Underwater backscattering causes significant high-frequency energy decay in the baseline (YOLOv11s). FinSight-Net effectively salvages these structural cues through EPA-FPN's long-range skip connections , ensuring superior fidelity and precise localization in turbid environments.}
%\vspace{-.2cm}
\label{fig: robust}
\end{figure}

\subsection{Visualization and Analysis }
To elucidate the internal representations learned by FinSight-Net, we perform a multifaceted qualitative analysis. This evaluation encompasses comparative detection results, frequency-domain energy assessments, and an interpretability study via adaptive branch activation weights and feature map visualizations. Together, these visualizations reveal how our physics-aware design restores structural fidelity and effectively decouples non-rigid fish targets from complex environmental noise.

\subsubsection{Comparative Detection Performance}
As illustrated in Fig \ref{fig:visualization}, we compare FinSight-Net with the YOLOv11s baseline under challenging conditions of high turbidity and dense occlusion. In turbid water, YOLOv11s suffers from significant missed detections due to the veiling effect of backscattering, which blurs morphological boundaries. In contrast, FinSight-Net localizes all targets with higher confidence, demonstrating the MS-DDSP bottleneck's ability to suppress noise and restore structural fidelity from degraded signals. Under dense occlusion, our model precisely delineates overlapping individuals and background targets where the baseline produces redundant or drifted boxes. This validates that the EPA-FPN's detail-filling mechanism successfully salvages critical high-frequency spatial cues typically filtered out during deep feature extraction, ensuring robust localization in complex aquaculture environments.

\subsubsection{Interpretability and Feature Representation}
\begin{figure}[pos=t]
    \centering
    \includegraphics[width=1\linewidth]{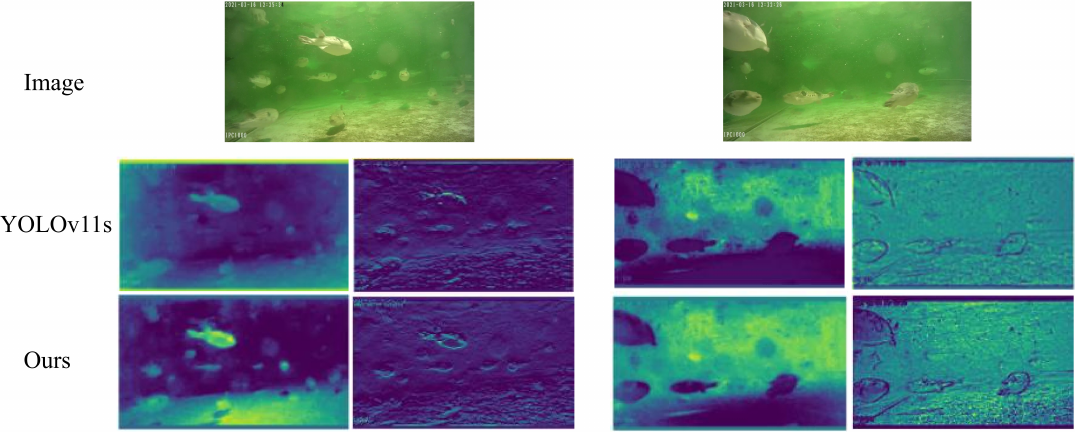}
\caption{Feature map visualization comparison. FinSight-Net enhances SNR via physics-aware MS-DDSP and EPA-FPN , effectively suppressing backscattering while salvaging high-frequency structural details to ensure precise fish-background discrimination in turbid environments.}
%\vspace{-.2cm}
\label{fig: Feature map}
\end{figure}

To further investigate the internal physics-aware mechanism, we analyze the model's interpretability and feature representations. As illustrated in the radar plot (Fig. \ref{fig: analysis}), the MS-DDSP bottleneck adaptively re-weights its parallel branches based on environmental conditions: Branch 2 (Decorrelation) is prioritized to suppress turbidity-induced backscattering, while Branch 3 (Spectral Weighting) is activated to compensate for wavelength-dependent absorption\citep{Akkaynak_2019_CVPR}. This dynamic resource allocation is qualitatively validated in the feature map comparison (Fig. \ref{fig: Feature map}). Compared to the YOLOv11s baseline, which retains significant background noise and blurred fish contours, FinSight-Net achieves a much higher Signal-to-Noise Ratio (SNR) at the feature level. By effectively filtering the veiling effect and salvaging fine-grained structural details—such as fin edges and body textures—our model ensures precise localization even in extreme aquaculture environments.

\subsection{Conclusion}
We propose FinSight-Net, a lightweight physics-aware detector for underwater aquaculture, explicitly addressing wavelength-dependent absorption and backscattering. FinSight-Net integrates an MS-DDSP bottleneck to decouple texture and contour cues for noise suppression and detail recovery, and an EPA-FPN to restore high-frequency spatial information via long-range skip connections with pruned fusion paths. Experiments on multiple benchmarks, especially UW-BlurredFish, show state-of-the-art performance: 92.8\% mAP, +4.8\% over YOLOv11s, with $\sim$29\% fewer parameters. Future work will explore adaptive frequency-domain interactions under varying turbidity/illumination and temporal consistency for non-rigid targets under occlusion.

\section{Acknowledgments} 
This work was supported by the National Natural Science Foundation of China under Grant (32573571, 62406052), the Key Research and Development Program of Liaoning Province (2023JH2/10200015), Natural Science Foundation of Liaoning Province under Grant 2024-BS-214, and the Basic Research Funding Projects of Liaoning Provincial Department of Education under Grant LJ212410158022.

\section*{Data availability}
Data will be made available on request.

\printcredits

\bibliographystyle{cas-model2-names}

\bibliography{cas-refs}

@ARTICLE{WANG2026111172,
  author  = {Jinfeng Wang and Jinze Lv and Zhipeng Cheng and Qiong Huang},
  title   = {FCNet: Accurate counting model of fish in complex underwater environments},
  journal = {Computers and Electronics in Agriculture}, 
  volume  = {240},
  year    = {2026},
  pages   = {111172}
}

@ARTICLE{JIA2026111141,
  author  = {Bo Jia and Xiaochan Wang and Yinyan Shi and Xiaolei Zhang and Zhen Xu and Jihao Wang and Haihui Yang and Dawei Qian},
  title   = {Efficient and lightweight model for detecting juvenile fish feeding behavior using feed pellet key point analysis},
  journal = {Computers and Electronics in Agriculture},
  volume  = {240},
  year    = {2026},
  pages   = {111141}
}

@ARTICLE{HE2026100896,
  author  = {Qingxuan He and Huihui Yu and Hanxiang Qin and Yupeng Mei and Ling Xu and Yingqian Chai and Cuili Li and Lihua Song and Daoliang Li and Yingyi Chen},
  title   = {Deep learning-based computer vision for fish behavior recognition in intensive aquaculture: A comprehensive review},
  journal = {Computer Science Review},
  volume  = {60},
  year    = {2026},
  pages   = {100896}
}

@InProceedings{Varghese_2025_CVPR,
    author    = {Varghese, Nisha and Rajagopalan, A. N.},
    title     = {Sea-ing in Low-light},
    booktitle = {Proceedings of the IEEE/CVF Conference on Computer Vision and Pattern Recognition (CVPR)},
    month     = {June},
    year      = {2025},
    pages     = {16629-16640}
}

@ARTICLE{ZHOU2022107372,
  author  = {Wen-Hui Zhou and Deng-Ming Zhu and Min Shi and Zhao-Xin Li and Ming Duan and Zhao-Qi Wang and Guo-Liang Zhao and Cheng-Dong Zheng},
  title   = {Deep images enhancement for turbid underwater images based on unsupervised learning},
  journal = {Computers and Electronics in Agriculture},
  volume  = {202},
  year    = {2022},
  pages   = {107372}
}

@ARTICLE{LI2025110764,
  author  = {Dashe Li and Siwei Zhao and Jingzhe Hu and Yufang Yang and Jinqiang Ding},
  title   = {An underwater image segmentation model for complex scenes in aquaculture using vision Transformer},
  journal = {Computers and Electronics in Agriculture}, 
  volume  = {238},
  year    = {2025},
  pages   = {110764}
}

@ARTICLE{WANG2026111248,
  author  = {Guangxu Wang and Yinfeng Hao and Daoliang Li},
  title   = {Multimodal estimation of fish weight via two-stream Transformer and adaptive RGB-D fusion},
  journal = {Computers and Electronics in Agriculture},
  volume  = {241},
  year    = {2026},
  pages   = {111248}
}

@ARTICLE{MUKSIT2022101847,
  author  = {Abdullah Al Muksit and Fakhrul Hasan and Fahad Hasan Bhuiyan Emon and Rakibul Haque and Arif Reza Anwary and Swakkhar Shatabda},
  title   = {YOLO-Fish: A robust fish detection model to detect fish in realistic underwater environment},
  journal = {Ecological Informatics},
  volume  = {72},
  year    = {2022},
  pages   = {101847}
}

@ARTICLE{YIN2025110612,
  author  = {Yihan Yin and Xueqian Sun and Guanghui Yu and Jiayi Wang and Daoliang Li and Yang Wang},
  title   = {CBFW-YOLOv8: Automated recognition method for fish body surface diseases in recirculating aquaculture systems},
  journal = {Computers and Electronics in Agriculture},
  volume  = {236},
  year    = {2025},
  pages   = {110612}
}

@ARTICLE{YU2023102108,
  author  = {Guoyan Yu and Ruilin Cai and Jinping Su and Mingxin Hou and Ruoling Deng},
  title   = {U-YOLOv7: A network for underwater organism detection},
  journal = {Ecological Informatics},
  volume  = {75},
  year    = {2023},
  pages   = {102108}
}

@ARTICLE{FENG2024102758,
  author  = {Jiangfan Feng and Tao Jin},
  title   = {CEH-YOLO: A composite enhanced YOLO-based model for underwater object detection},
  journal = {Ecological Informatics},
  volume  = {82},
  year    = {2024},
  pages   = {102758}
}

@InProceedings{Liu_2018_CVPR,
author = {Liu, Shu and Qi, Lu and Qin, Haifang and Shi, Jianping and Jia, Jiaya},
title = {Path Aggregation Network for Instance Segmentation},
booktitle = {Proceedings of the IEEE Conference on Computer Vision and Pattern Recognition (CVPR)},
month = {June},
year = {2018}
}

@InProceedings{Ghiasi_2019_CVPR,
author = {Ghiasi, Golnaz and Lin, Tsung-Yi and Le, Quoc V.},
title = {NAS-FPN: Learning Scalable Feature Pyramid Architecture for Object Detection},
booktitle = {Proceedings of the IEEE/CVF Conference on Computer Vision and Pattern Recognition (CVPR)},
month = {June},
year = {2019}
}

@InProceedings{Tan_2020_CVPR,
author = {Tan, Mingxing and Pang, Ruoming and Le, Quoc V.},
title = {EfficientDet: Scalable and Efficient Object Detection},
booktitle = {Proceedings of the IEEE/CVF Conference on Computer Vision and Pattern Recognition (CVPR)},
month = {June},
year = {2020}
}

@INPROCEEDINGS{10.1117/12.3022812,
  author    = {Mingjian Zhu},
  title     = {Dynamic feature pyramid networks for object detection},
  booktitle = {Fifteenth International Conference on Signal Processing Systems (ICSPS 2023)},
  volume    = {13091},
  year      = {2024},
  pages     = {130911N}
}

@ARTICLE{XU2023204,
  author  = {Shubo Xu and Minghua Zhang and Wei Song and Haibin Mei and Qi He and Antonio Liotta},
  title   = {A systematic review and analysis of deep learning-based underwater object detection},
  journal = {Neurocomputing},
  volume  = {527},
  year    = {2023},
  pages   = {204-232}
}

@ARTICLE{ZHOU2024102680,
  author  = {Hui Zhou and Meiwei Kong and Hexiang Yuan and Yanyan Pan and Xinru Wang and Rong Chen and Weiheng Lu and Ruizhi Wang and Qunhui Yang},
  title   = {Real-time underwater object detection technology for complex underwater environments based on deep learning},
  journal = {Ecological Informatics},
  volume  = {82},
  year    = {2024},
  pages   = {102680}
}

@ARTICLE{7485869,
  author  = {Shaoqing Ren and Kaiming He and Ross Girshick and Jian Sun},
  title   = {Faster R-CNN: Towards Real-Time Object Detection with Region Proposal Networks},
  journal = {IEEE Transactions on Pattern Analysis and Machine Intelligence},
  volume  = {39},
  year    = {2017},
  pages   = {1137-1149}
}

@ARTICLE{YI2024127488,
  author  = {Dewei Yi and Hasan Bayarov Ahmedov and Shouyong Jiang and Yiren Li and Sean Joseph Flinn and Paul G. Fernandes},
  title   = {Coordinate-Aware Mask R-CNN with Group Normalization: A underwater marine animal instance segmentation framework},
  journal = {Neurocomputing},
  volume  = {583},
  year    = {2024},
  pages   = {127488}
}

@article{shen2024imagpose,
  title={Imagpose: A unified conditional framework for pose-guided person generation},
  author={Shen, Fei and Tang, Jinhui},
  journal={Advances in neural information processing systems},
  volume={37},
  pages={6246--6266},
  year={2024}
}

@inproceedings{shen2025imagdressing,
  title={Imagdressing-v1: Customizable virtual dressing},
  author={Shen, Fei and Jiang, Xin and He, Xin and Ye, Hu and Wang, Cong and Du, Xiaoyu and Li, Zechao and Tang, Jinhui},
  booktitle={Proceedings of the AAAI Conference on Artificial Intelligence},
  volume={39},
  number={7},
  pages={6795--6804},
  year={2025}
}

@inproceedings{shen2024advancing,
title={Advancing Pose-Guided Image Synthesis with Progressive Conditional Diffusion Models},
author={Fei Shen and Hu Ye and Jun Zhang and Cong Wang and Xiao Han and Yang Wei},
booktitle={The Twelfth International Conference on Learning Representations},
year={2024},
url={https://openreview.net/forum?id=rHzapPnCgT}
}

@INPROCEEDINGS{pmlr-v162-dimitrakopoulos22a,
  author    = {Panagiotis Dimitrakopoulos and Giorgos Sfikas and Christophoros Nikou},
  title     = {Variational Feature Pyramid Networks},
  booktitle = {Proceedings of the 39th International Conference on Machine Learning},
  volume    = {162},
  year      = {2022},
  pages     = {5142--5152}
}

@INPROCEEDINGS{Wang2023GoldYOLO,
  author    = {Chengqiang Wang and Wenrui Dai and Kai Han and Yunhe Wang and Jidong Wang and Zhipeng Hu and Bin Xiao and Junli Gu},
  title     = {Gold-YOLO: Efficient Object Detector with Gather-and-Distribute Mechanism},
  booktitle = {Advances in Neural Information Processing Systems (NeurIPS 2023)},
  volume    = {36},
  year      = {2023},
  pages     = {47249--47261}
}

@ARTICLE{10852283,
  author  = {Mahmoud Elmezain and Lyes Saad Saoud and Atif Sultan and Mohamed Heshmat and Lakmal Seneviratne and Irfan Hussain},
  title   = {Advancing Underwater Vision: A Survey of Deep Learning Models for Underwater Object Recognition and Tracking},
  journal = {IEEE Access},
  volume  = {13},
  year    = {2025},
  pages   = {17830-17867}
}

@InProceedings{Chollet_2017_CVPR,
author = {Chollet, Francois},
title = {Xception: Deep Learning With Depthwise Separable Convolutions},
booktitle = {Proceedings of the IEEE Conference on Computer Vision and Pattern Recognition (CVPR)},
month = {July},
year = {2017}
}

@InProceedings{Hua_2018_CVPR,
author = {Hua, Binh-Son and Tran, Minh-Khoi and Yeung, Sai-Kit},
title = {Pointwise Convolutional Neural Networks},
booktitle = {Proceedings of the IEEE Conference on Computer Vision and Pattern Recognition (CVPR)},
month = {June},
year = {2018}
}

@INPROCEEDINGS{10.1117/12.3094538,
  author    = {Junyi Shao},
  title     = {Underwater image modeling and enhancement method based on the Jaffe–McGlamery optical imaging model},
  booktitle = {Second International Conference on Communication, Information, and Digital Technologies (CIDT 2025)},
  volume    = {14064},
  year      = {2026},
  pages     = {140640N}
}

@InProceedings{Zhao_2024_CVPR1,
    author    = {Zhao, Chen and Cai, Weiling and Dong, Chenyu and Hu, Chengwei},
    title     = {Wavelet-based Fourier Information Interaction with Frequency Diffusion Adjustment for Underwater Image Restoration},
    booktitle = {Proceedings of the IEEE/CVF Conference on Computer Vision and Pattern Recognition (CVPR)},
    month     = {June},
    year      = {2024},
    pages     = {8281-8291}
}

@InProceedings{Wang_2020_CVPR_Workshops,
author = {Wang, Chien-Yao and Liao, Hong-Yuan Mark and Wu, Yueh-Hua and Chen, Ping-Yang and Hsieh, Jun-Wei and Yeh, I-Hau},
title = {CSPNet: A New Backbone That Can Enhance Learning Capability of CNN},
booktitle = {Proceedings of the IEEE/CVF Conference on Computer Vision and Pattern Recognition (CVPR) Workshops},
month = {June},
year = {2020}
}

@article{shen2025imagharmony,
  title={IMAGHarmony: Controllable Image Editing with Consistent Object Quantity and Layout},
  author={Shen, Fei and Du, Xiaoyu and Gao, Yutong and Yu, Jian and Cao, Yushe and Lei, Xing and Tang, Jinhui},
  journal={arXiv preprint arXiv:2506.01949},
  year={2025}
}

@article{shen2025imaggarment,
  title={IMAGGarment-1: Fine-Grained Garment Generation for Controllable Fashion Design},
  author={Shen, Fei and Yu, Jian and Wang, Cong and Jiang, Xin and Du, Xiaoyu and Tang, Jinhui},
  journal={arXiv preprint arXiv:2504.13176},
  year={2025}
}

@inproceedings{shenlong,
  title={Long-Term TalkingFace Generation via Motion-Prior Conditional Diffusion Model},
  author={Shen, Fei and Wang, Cong and Gao, Junyao and Guo, Qin and Dang, Jisheng and Tang, Jinhui and Chua, Tat-Seng},
  booktitle={Forty-second International Conference on Machine Learning}
}

@ARTICLE{QIN201649,
  author  = {Hongwei Qin and Xiu Li and Jian Liang and Yigang Peng and Changshui Zhang},
  title   = {DeepFish: Accurate underwater live fish recognition with a deep architecture},
  journal = {Neurocomputing},
  volume  = {187},
  year    = {2016},
  pages   = {49-58}
}

@INPROCEEDINGS{NEURIPS2019_bdbca288,
  author    = {Adam Paszke and Sam Gross and Francisco Massa and Adam Lerer and James Bradbury and Gregory Chanan and Trevor Killeen and Zeming Lin and Natalia Gimelshein and Luca Antiga and Alban Desmaison and Andreas Kopf and Edward Yang and Zachary DeVito and Martin Raison and Alykhan Tejani and Sasank Chilamkurthy and Benoit Steiner and Lu Fang and Junjie Bai and Soumith Chintala},
  title     = {PyTorch: An Imperative Style, High-Performance Deep Learning Library},
  booktitle = {Advances in Neural Information Processing Systems},
  volume    = {32},
  year      = {2019},
  pages     = {8024--8035}
}

@ARTICLE{d8d62392-9a37-31e7-ad3b-37a6f6ee8ef6,
  author  = {Herbert Robbins and Sutton Monro},
  title   = {A Stochastic Approximation Method},
  journal = {The Annals of Mathematical Statistics},
  volume  = {22},
  year    = {1951},
  pages   = {400--407}
}

@software{yolov8_ultralytics,
  author       = {Glenn Jocher and Ayush Chaurasia and Jing Qiu},
  title        = {Ultralytics YOLOv8},
  version      = {8.0.0},
  year         = {2023},
  url          = {https://github.com/ultralytics/ultralytics},
  publisher    = {GitHub}
}

@INPROCEEDINGS{NEURIPS2024_c34ddd05,
  author    = {Ao Wang and Hui Chen and Lihao Liu and Kai Chen and Zijia Lin and Jungong Han and Guiguang Ding},
  title     = {YOLOv10: Real-Time End-to-End Object Detection},
  booktitle = {Advances in Neural Information Processing Systems},
  volume    = {37},
  year      = {2024},
  pages     = {107984--108011}
}

@software{yolov11_ultralytics,
  author = {Glenn Jocher and Jing Qiu},
  title = {Ultralytics YOLO11},
  version = {11.0.0},
  year = {2024},
  url = {https://github.com/ultralytics/ultralytics}
}

@INPROCEEDINGS{Zhao_2024_CVPR2,
  author    = {Yian Zhao and Wenyu Lv and Shangliang Xu and Jinman Wei and Guanzhong Wang and Qingqing Dang and Yi Liu and Jie Chen},
  title     = {DETRs Beat YOLOs on Real-time Object Detection},
  booktitle = {Proceedings of the IEEE/CVF Conference on Computer Vision and Pattern Recognition (CVPR)},
  year      = {2024},
  pages     = {16965-16974}
}

@InProceedings{Akkaynak_2019_CVPR,
author = {Akkaynak, Derya and Treibitz, Tali},
title = {Sea-Thru: A Method for Removing Water From Underwater Images},
booktitle = {Proceedings of the IEEE/CVF Conference on Computer Vision and Pattern Recognition (CVPR)},
month = {June},
year = {2019}
}

@ARTICLE{ZHANG2026111227,
  author  = {Qinyue Zhang and Shasha Wang and Tianshu Zhang and Guiming Ren and Lingling Zhang and Yangfan Wang and Bing Zheng and Juan Li and Haiyong Zheng},
  title   = {Toward smart aquaculture: A review of multimodal methods, datasets, and applications from the modality perspective},
  journal = {Computers and Electronics in Agriculture},
  volume  = {240},
  year    = {2026},
  pages   = {111227}
}


\bio{}

\endbio

%\bio{figs/pic1}
\endbio

\end{document}